\documentclass[runningheads]{llncs}
\usepackage{amssymb}
\usepackage{footmisc}
\usepackage[T1]{fontenc}
\usepackage{graphicx}
\usepackage{algpseudocode}
\usepackage{algorithm}
\newcommand{\eq}[1]{\begin{small}\begin{align}#1\end{align}\end{small}}

\usepackage{amsmath}
\DeclareMathOperator*{\argmax}{arg\,max} 

\begin{document}
\title{Improving Next Tokens via Second-to-Last Predictions with Generate and Refine}

\author{Johannes Schneider\inst{1}}
\institute{Department of Computer Science and Information Systems\\ University of Liechtenstein, Vaduz, Liechtenstein  \\ \email{johannes.schneider@uni.li}}  

%
\maketitle              

\begin{abstract}
Autoregressive language models like GPT aim to predict next tokens, while autoencoding models such as BERT are trained on tasks such as predicting masked tokens. We train a decoder-only architecture for predicting the second to last token for a sequence of tokens. Our approach yields higher computational training efficiency than BERT-style models by employing a structured deterministic approach to masking tokens. We use our model to improve the next token predictions of a standard GPT by combining both predictions in a ``generate-then-refine'' approach. We demonstrate on different variants of GPT-2 and different datasets that (not unexpectedly) second to last token predictions are much more accurate, i.e., more than 15\% higher accuracy than standard next token predictions. The ``generate-then-refine'' approach also demonstrates notable improvements in next-token predictions, yielding smaller yet consistent and significant gains. 
\end{abstract}

\section{Introduction}
\begin{figure}
\centering
\fbox{%
\includegraphics[width=0.9\textwidth]{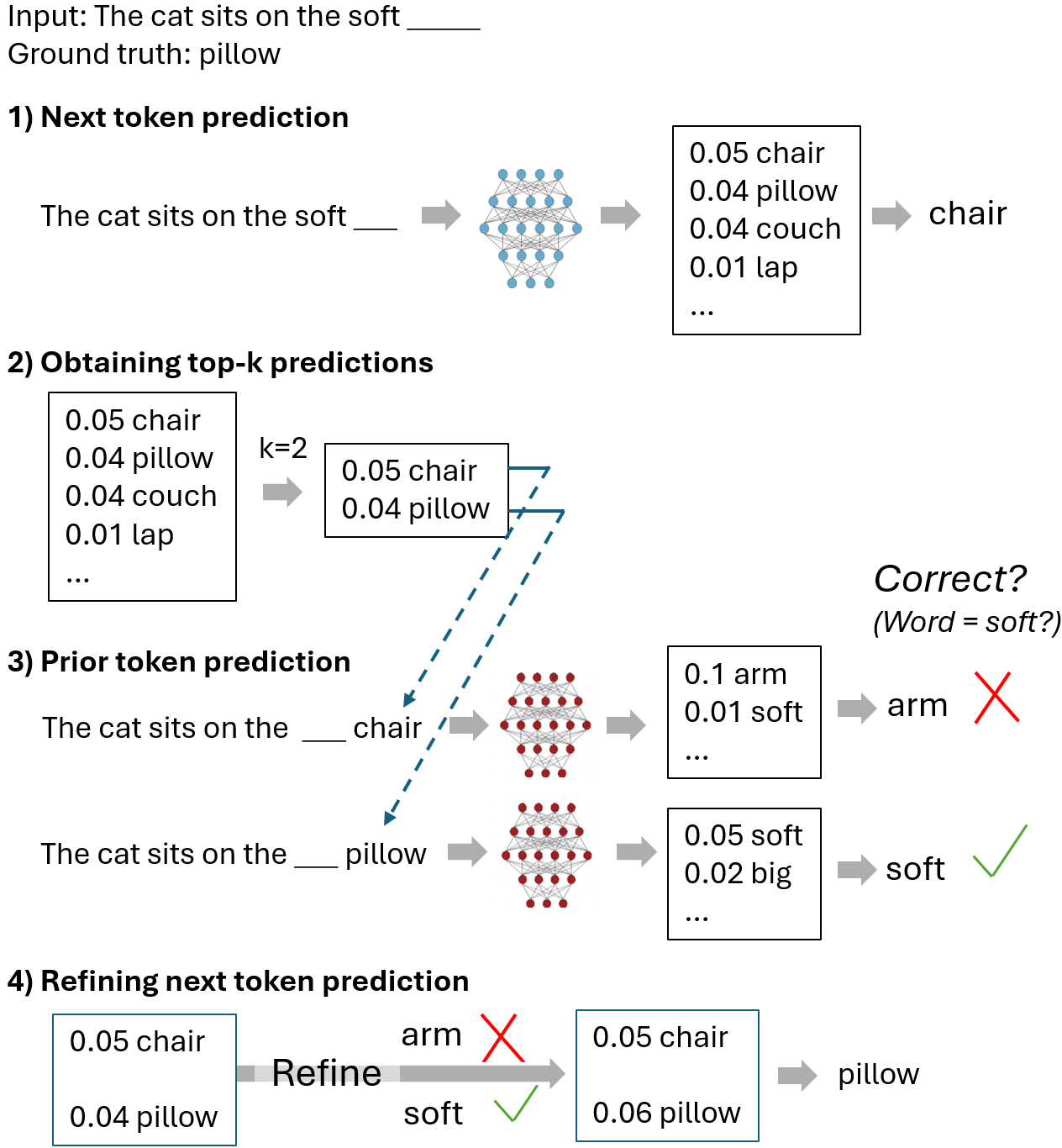}}
\caption{Conceptual outline of our ``generate-then-refine'' approach using second-to-last token prediction formalized in Algorithm \ref{alg:comb}}\label{fig:concept}
\end{figure}

Large language models based on transformers~\cite{Scale17} have disrupted the field of natural language processing. In particular, autoregressive models such as the Generative Pre-trained Transformer (GPT) series, introduced in 2018~\cite{rad18}, marked a major step forward due to their ability to generate mostly coherent and contextually relevant text by predicting the next token solely based on preceding tokens. Conversely, autoencoding models like BERT (Bidirectional Encoder Representations from Transformers)~\cite{BERT18} led to large improvements in understanding and representation tasks. They predict masked tokens within a sequence, thereby capturing both preceding and succeeding context.
Efforts to bridge the gap between these models have led to innovations such as XLNet~\cite{XLNet19} integrating permutation-based training to capture bidirectional context while maintaining autoregressive properties. We pursue a different avenue to improve next token predictions by adopting a ``generate-then-refine'' approach. In our approach, a standard autoregressive model predicts the top-k next tokens, which are provided as context, i.e., last token, to an auto-encoding model aiming to predict the second-to-last token (see Figure \ref{fig:concept}). Thereby, we incorporate bidirectional context into a unidirectional model. Predictions of both models are then combined to improve next-token predictions. That is, we use prior token predictions as a form of feedback (or verification) to improve reliability and accuracy of next token predictions in standard autoregressive models. We believe that this approach is of great interest as the ``generate-then-refine'' has received considerable attention in recent years~\cite{pan24aut}. In particular, on a larger scale, our work contributes to the ongoing debate on when LLMs can self-correct during inference, where self-correction might also be triggered by external sources~\cite{kam24can}. That is, we provide a novel approach for self-correction. 

Our implementation is also a novel way to remedy shortcomings of earlier works such as~\cite{BERT18}, e.g., eliminating the need for a dedicated mask token, and thereby increasing training efficiency due to a structured, deterministic choice of masking locations and permutation of input sequences. 

We evaluate our approach on three variants of GPT-2~\cite{GPT-219} across three datasets, showing that (as expected) leveraging bidirectional context leads to much higher prediction accuracy than next token predictions. However, translating high accuracy for second-to-last token predictions into substantial gains for next token predictions is more challenging, i.e., while we found consistent and statistically significant gains for our vanilla model using the same parameters across all datasets, the gains are small. Nevertheless, in the light of the challenges related to self-correction~\cite{kam24can}, we believe that our novel approach is of considerable interest. We also see it as a first step that can be further improved.

\section{Methodology}
Our approach, illustrated in Figure \ref{fig:concept}, employs two models: A standard, unidirectional autoregressive model $f_{n}$ for next token prediction, i.e., in our evaluation we use GPT-2 variants, and a second, bidirectional (autoencoding) model $f_s$ with the same architecture trained to predict the second-to-last token. The latter model serves as an evaluator or refiner of the top-k predictions of the autoregressive model $f_n$ by predicting the second to last token given each of the top-k predictions as last token. Our method increases the likelihood of all top-k predictions for which the second to last token was predicted correctly.

More formally, the autoregressive language model $f_n$ computes a probability distribution $p_n$ aiming to approximate the true conditional probability distribution $p^*$ of a token $y_t$ given prior tokens $y_{i<t}$:
\eq{
f_n(y_t|y_{i<t})=p_n(y_t|y_{0},y_1,...,y_{t-1})\\\approx p^*(y_t|y_{0},y_1,...,y_{t-1})
}
Analogously, the autoencoding model predicts the second-to-last token $y_{t-1}$ given tokens $y_{i<(t-1)}$ and the last token $y_t$:
\eq{
f_s(y_{t-1}|y_{i<t-1},y_t)=p_s(y_{t-1}|y_{0},y_1,...,y_{t-3},y_{t-2},y_t)\\ \approx p^*(y_{t-1}|y_{0},y_1,...,y_{t-3},y_{t-2},y_t)
} 
To refine the next token prediction $y$ of $f_n$, we compute the top-k most likely tokens $Top_k$ (in Algorithm \ref{alg:comb}). For each token $y \in Top_k$, we compute the output of the autoencoding model $f_s$, i.e., $f_s(x|y_{i<t-1},y)$. If the most likely token predicted by $f_s$ for token $y$ is indeed the correct second to last token $y_{t-1}$, i.e., $y_{t-1}=\argmax_x f_s(x|y_{i<t-1},y)$, we multiply the probability $f_n(y|y_{i<t})$ by a factor $(1+w)$, where $w$ is a parameter.  We found this to work somewhat better than other approaches such as adding a fixed value. However, the essential point is that a next token prediction $y$ of model $f_n$ is seen as more likely correct if the assessor model $f_s$ correctly identifies the second to last token using the prediction $y$ as last token. We discuss the underlying assumptions of our method in Section \ref{sec:dis}, underpinning them with empirical outcomes.
For implementation, it is more efficient to work on logits to avoid softmax computations yielding probabilities. Code is at \emph{\url{https://github.com/JohnTailor/AutoEncGenerateThenRefine/}}.

\begin{algorithm}
\caption{Algorithm AGR (Auto-encoding ``Generate-then-Refine'')} \label{alg:comb}
\begin{algorithmic}[1] 
\State \textbf{Input:} Models $f_n$, $f_s$ and tokens $y_{i<t}$ 
\State \textbf{Output:} Next token prediction $y_t$
\State $k:=15$ \Comment{number of tokens to consider}
\State $w:=0.05$ \Comment{(incremental) factor if second-to-last tokens is correct}
\State $Top_k:= \{ $Tokens $y_t | y_t \text{ among $k$ tokens with max probability }  f_n(y|y_{i<t})\}$
\State $y'_t:= \argmax_{y \in Top_k} p_n(y|y_{i<t})\cdot (1 + w\cdot \textbf{1}_{\{y_{t-1}=\argmax_x f_s(x|y_{i<t-1},y)\}})$ 
\State Return $y'_t$
\end{algorithmic}
\end{algorithm}

\subsection{Implementation for efficient training} \label{sec:imp}
\begin{figure}
\centering
\fbox{%
\includegraphics[width=0.9\textwidth]{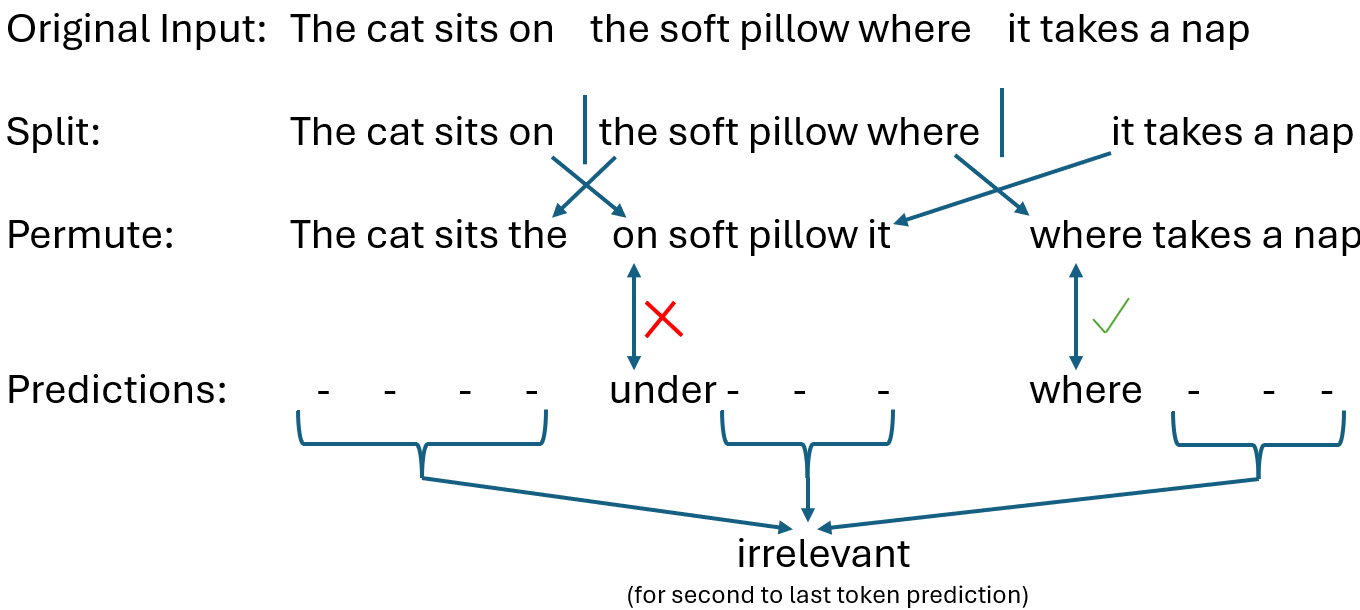}}
\caption{Example of how input sequences are processed to obtain sequences for training the second to last token prediction model $f_s$}\label{fig:per}
\end{figure}

Previous works like BERT~\cite{BERT18} commonly use a special token to indicate a masked token, typically masking only a small fraction of words (e.g. around 10\%) for a given text, making autoencoding slow to train. We forgo the need for a mask token and increase the number of predicted tokens. We do so by eliminating randomness in choices of mask locations in the input. We decompose the input sequences into short subsequences and permute them so that they better fit existing frameworks for next token prediction. That is, the decoder of a transformer~\cite{Scale17} employs a causal attention mask that ensures that for a given input $S=(y_0,y_1,...,y_{t-1})$ of $t$ tokens,  to predict token $y_i$ the decoder only sees tokens $y_{j<i}$, i.e., it cannot see the future tokens it should predict. This mechanism ensures that for training with an input sequence of length $t$ we also obtain $t$ predictions to compute the loss. However, our model $f_s$ (implemented as a decoder) can effectively see the `next' token but not the preceding one, which makes masking challenging. A simple strategy to overcome this issue would be to train without any causal attention mask, altering the input sequence a bit by removing the token to predict $y_{t-1}$, i.e., $S=(y_0,y_1,...,y_{t-2},y_t)$. The transformer predicts just a single token $y_{t-1}$ for the entire sequence (technically, we can simply ignore all predictions except the last one for loss computation). The problem with this approach is that it slows down training by a factor of $t$, which is more than 100 times slower in practice, i.e., typical input lengths are significantly larger than 100 tokens. Thus, this approach is impractical. To improve training speed, we introduce a compromise. We split the entire input sequence into subsequences of length $l\ll t$, i.e., we set $l=4$. For each of these subsequences, we obtain a prediction that is used for loss computation, effectively slowing down training by factor of $l\ll t$. More precisely, for training of $f_s$ we replace the first subsequence of $l=4$ tokens $(y_0,y_1,y_2,y_3)$ by $(y_0,y_1,y_2,y_4)$. More generally, we move tokens at positions $j\cdot l+(l-1)$ to position $j\cdot l+l$ leading to a permuted sequence like $S'=(y_0,y_1,y_2,y_4 ,y_3,y_5,y_6,y_8, y_7,y_9,y_{10},y_{12},...)$. We only use the predictions for tokens $j\cdot l+l$ for loss computations. The process for altering the original input sequence for training the model for second-to-last token prediction $f_s$ is illustrated in Figure \ref{fig:per}.\footnote{An alternative approach is to use a custom causal mask, which is more tricky to implement.}

\section{Evaluation}
We evaluated Algorithm (\ref{alg:comb}) AGR using three datasets and three variants of the GPT-2 model~\cite{GPT-219} differing mostly in terms of size, i.e., the number of parameters. Note that GPT-2 is very similar to newer models such as GPT-3 and GPT-4 (see~\cite{sch24comes} for a more detailed comparison). More specifically, we built on the NanoGPT repository by Karpathy\footnote{\url{https://github.com/karpathy/nanoGPT} \label{foot:nano}}. We computed loss metrics and prediction accuracy but focused our discussion on accuracy for simplicity, as the two are highly correlated.

\subsection{Datasets, Models and Training} 
We use three text datasets that vary in text type and size (see Table \ref{tab:dat}). Each dataset was split randomly into 90\% training and 10\% validation. For each dataset, we train two models, i.e., a next token prediction model $f_n$ and prior token prediction model $f_s$ with key parameters stated in Table \ref{tab:mod}. For data preprocessing such as tokenization model definitions and training, we leveraged the NanoGPT repository code\footref{foot:nano}. That is, we performed warmup iterations (Column 'WarmupIt.' in Table \ref{tab:mod}) with the minimum learning rate set to 1/10 of the starting learning rate of (see Column 'LearningRate' in Table \ref{tab:mod}) and conducted a fixed number of iterations given by column '\#Iters' using a batch size of 768, with each entry having `BlockSize' many tokens. The initial learning rate 'LearningRate' after the warmup phase is reduced until it reaches 1/10 of the initial rate at the end of training. The model architecture hyperparameters are given by the number of layers (\#Layer), number of attention heads (\#Head) and embedding dimension (EmbDim) yielding a total number of parameters given by '\#Para.'.\footnote{Note that the number of parameters in our model might be smaller than those for a native GPT-2 model as we removed tokens that did not occur in the corpus and remapped the others to reduce the embedding matrix.}
We trained each model in Table \ref{tab:mod} once. For evaluation of our Algorithm \ref{alg:comb} we  performed 70000 randomly chosen token predictions for  both training and validation data. Each predicted token is based on an input of maximal length, i.e., of 'BlockSize' length (Table \ref{tab:mod}). We performed ten evaluation runs and reported the mean and standard deviation of accuracies of both models $f_n$ and $f_s$ and Algorithm AGR (\ref{alg:comb}).\footnote{As we chose data randomly, some data might overlap across runs, but as the validation data size of the smallest dataset simplewiki is more than a factor of 20 larger than the evaluation data, the overlap is expected to be limited.}

To ensure a fair comparison in terms of the amount of training, for next token prediction we also only use tokens $j\cdot l+l$ for loss computations as for second-to-last token prediction (see Figure \ref{fig:per} described in Section \ref{sec:imp}). However, we found that significantly longer training of the next token prediction model yields only limited gains without additional measures such as increasing model capacity and training data.

\noindent\textbf{Computational Settings:} Our experiments were executed on an Ubuntu 22.04 system equipped with Python 3.12, PyTorch 2.5, CUDA 12.5, running on a server with 512 GB of RAM, a 64-core AMD EPYC 9554 CPU, and two NVIDIA H100 NVL GPUs.

\begin{table}
\caption{Datasets}\label{tab:dat}
\scriptsize {
\begin{tabular}{|l|c|c|}
\hline
\textbf{Dataset} &  \textbf{\#Tokens} & \textbf{Description / URL}   \\ \hline
simplewiki & 43 Millions & Simple Wikipedia \\
& & \url{https://dumps.wikimedia.org/simplewiki/latest/} \\
&& \url{simplewiki-latest-pages-articles.xml.bz2} \footnote{Downloaded in Aug. 2024}\\ \hline
openwebtext & 1.7 Billion & Open-source replication of the WebText dataset from OpenAI \\
& & \url{http://Skylion007.github.io/OpenWebTextCorpus} \\ \hline
Stories & 364 Millions & Children stories generated by Llama3 8B \\ 
& & \url{https://huggingface.co/datasets/Seppel123/Stories\_by\_Llama3} \\
\hline
\end{tabular}
}
\end{table}

\begin{table}
\caption{Model configuration for each dataset}\label{tab:mod}
\scriptsize {
\begin{tabular}{|l||c|c|c|c|c| c|c|c|}
\hline
\textbf{Dataset} & \multicolumn{8}{c|}{\textbf{Model Configuration}}  \\ \cline{2-9}
& {\#Para.} &{LearningRate} &  {WarmupIt.} & {\#Iters} & {BlockSize} & {\#Layer} & {\#Head} & {EmbDim} \\ \hline
openwebtext & 800 Mio & 7e-4 & 4000 & 400000 & 208 & 30 & 16 & 1424 \\
simplewiki & 227 Mio & 1e-3 & 2000 & 200000 & 112 & 14 & 4 & 1024 \\
Stories & 300 Mio & 1e-3 & 2000 & 400000 & 144 & 16 & 8 & 1152 \\
\hline
\end{tabular}
}
\end{table}

\subsection{Results}
The results in Table \ref{tab:res} show significant gains for dataset SimpleWiki but rather small gains for other datasets. We conducted a t-test  for the vanilla configuration ($k=15$ and $w=0.05$) and found that gains using Algorithm AGR (\ref{alg:comb}), i.e., $\Delta_{AGR-f_n}:=Acc^{va}_{AGR}-Acc^{va}_s$, are greater than zero (p-value <0.05) for each dataset. Thus, while our approach yields gains across all datasets, considering the fact that the model accuracies $f_s$ for second-to-last token prediction are much larger than those of the next token prediction model (i.e., differing by about 20\% on validation accuracy), the gains ranging from 0.03\% to 0.22\% for the vanilla configuration appear small. A deeper analysis identified several cases where the refiner model $f_s$ is not of much help. First, if the second-to-last or next token is very common, e.g., a word like ``a''. In this case, we found, for example, that the model $f_s$ might predict the correct token for multiple of the top $k$ tokens of $f_n$, e.g., if the second to last token is `a' and among the top k next predicted tokens are nouns like 'house' or 'car', the token `a' might be predicted for both nouns.
Second, if the token is rather uncommon and both models fail, i.e., the correct token is not among the top-$k$ tokens. For instance, we found that the correct token is not among the top $k$ in 50\% of all predictions for $k=2$ for the simplewiki dataset and in 75\% for $k=15$ for the openwebtext dataset. Furthermore, on this dataset, the model $f_s$ only achieves an accuracy of about 60\%  meaning that its predictions are noisy. Also, we found that, if $f_s$ is not performing well (e.g.,  achieving about 60\%), relying too much on the prediction of $f_s$, i.e., choosing a large weight $w$ ($w=0.1$), leads to lower accuracy than not using Algorithm AGR \ref{alg:comb}, indicating that reliance on the model $f_s$ should be limited despite its higher accuracies compared to $f_n$. However, if second to last token prediction $f_s$ is fairly reliably it appears that larger weights such as $w=0.1$ clearly outperform lower weights -- A t-test shows that it performs better than any other configuration with p<0.001 for simplewiki. To provide evidence in favor of the conjecture that higher validation accuracy of the validation model leads to larger weights $w$ being more favorable, we loaded a checkpoint of our model trained only for 20\% of all iterations stated in Table \ref{tab:mod}. For the model from the checkpoint the validation accuracy was only $64\%$ for the refiner model $f_s$. In turn choosing $w=0.1$ was significantly worse (p<0.05) than the best configuration.

\begin{table}
\caption{Results: Accuracies for train(tr) and validation(va) data for models $f_s$, $f_n$ and Algorithm AGR varying parameters $k$ and $w$; $\Delta_{AGR-f_n}>0$ indicates gains of Algorithm AGR, which hold in all settings except for large $w$}\label{tab:res}
\scriptsize {
\begin{tabular}{|l|l|l||l|l|l|| l|l|l|l|}
\hline
Dataset & k & w & $Acc^{tr}_{n}$ & $Acc^{tr}_{s}$ & $Acc^{tr}_{AGR}$ & $Acc^{va}_{n}$ & $Acc^{va}_{s}$ & $Acc^{va}_{AGR}$ & \textbf{$\Delta_{AGR-f_n}$}\\ \hline
Stories& 15& 0.05 & 77.82\tiny{\text{$\pm$}0.23}&93.08\tiny{\text{$\pm$}0.18}&78.08\tiny{\text{$\pm$}0.24}&65.15\tiny{\text{$\pm$}0.25}&83.96\tiny{\text{$\pm$}0.12}&65.2\tiny{\text{$\pm$}0.25}&0.04\tiny{\text{$\pm$}0.02}\\
Stories& 5& 0.05 & 77.84\tiny{\text{$\pm$}0.39}&92.97\tiny{\text{$\pm$}0.14}&78.1\tiny{\text{$\pm$}0.4}&65.09\tiny{\text{$\pm$}0.15}&84.01\tiny{\text{$\pm$}0.15}&65.11\tiny{\text{$\pm$}0.16}&0.03\tiny{\text{$\pm$}0.03}\\
Stories& 2& 0.05 & 77.84\tiny{\text{$\pm$}0.4}&93.08\tiny{\text{$\pm$}0.18}&78.09\tiny{\text{$\pm$}0.39}&65.16\tiny{\text{$\pm$}0.12}&83.94\tiny{\text{$\pm$}0.12}&65.18\tiny{\text{$\pm$}0.12}&0.02\tiny{\text{$\pm$}0.03}\\
Stories& 5& 0.01 & 77.86\tiny{\text{$\pm$}0.32}&93.15\tiny{\text{$\pm$}0.12}&77.93\tiny{\text{$\pm$}0.32}&65.11\tiny{\text{$\pm$}0.15}&83.94\tiny{\text{$\pm$}0.12}&65.13\tiny{\text{$\pm$}0.15}&0.02\tiny{\text{$\pm$}0.01}\\
Stories& 5& 0.1 & 77.82\tiny{\text{$\pm$}0.27}&93.06\tiny{\text{$\pm$}0.12}&78.14\tiny{\text{$\pm$}0.28}&65.05\tiny{\text{$\pm$}0.21}&83.93\tiny{\text{$\pm$}0.15}&64.99\tiny{\text{$\pm$}0.21}&-0.06\tiny{\text{$\pm$}0.04}\\ \hline
openwebtext& 15& 0.05 & 42.55\tiny{\text{$\pm$}0.3}&63.06\tiny{\text{$\pm$}0.22}&42.6\tiny{\text{$\pm$}0.32}&42.4\tiny{\text{$\pm$}0.19}&62.93\tiny{\text{$\pm$}0.2}&42.46\tiny{\text{$\pm$}0.19}&0.06\tiny{\text{$\pm$}0.04}\\
openwebtext& 5& 0.05 & 42.47\tiny{\text{$\pm$}0.32}&62.95\tiny{\text{$\pm$}0.44}&42.52\tiny{\text{$\pm$}0.34}&42.45\tiny{\text{$\pm$}0.17}&63.01\tiny{\text{$\pm$}0.19}&42.5\tiny{\text{$\pm$}0.17}&0.05\tiny{\text{$\pm$}0.02}\\
openwebtext& 2& 0.05 & 42.68\tiny{\text{$\pm$}0.29}&63.07\tiny{\text{$\pm$}0.24}&42.69\tiny{\text{$\pm$}0.27}&42.54\tiny{\text{$\pm$}0.22}&62.95\tiny{\text{$\pm$}0.12}&42.58\tiny{\text{$\pm$}0.22}&0.03\tiny{\text{$\pm$}0.02}\\
openwebtext& 5& 0.01 & 42.59\tiny{\text{$\pm$}0.26}&62.9\tiny{\text{$\pm$}0.27}&42.62\tiny{\text{$\pm$}0.25}&42.47\tiny{\text{$\pm$}0.09}&63.02\tiny{\text{$\pm$}0.24}&42.49\tiny{\text{$\pm$}0.09}&0.02\tiny{\text{$\pm$}0.01}\\
openwebtext& 5& 0.1 & 42.54\tiny{\text{$\pm$}0.28}&63.06\tiny{\text{$\pm$}0.39}&42.53\tiny{\text{$\pm$}0.3}&42.56\tiny{\text{$\pm$}0.16}&62.88\tiny{\text{$\pm$}0.17}&42.54\tiny{\text{$\pm$}0.15}&-0.02\tiny{\text{$\pm$}0.04}\\ \hline

simplewiki& 5& 0.1 & 95.9\tiny{\text{$\pm$}0.14}&98.7\tiny{\text{$\pm$}0.08}&96.57\tiny{\text{$\pm$}0.14}&43.44\tiny{\text{$\pm$}0.18}&63.8\tiny{\text{$\pm$}0.16}&43.78\tiny{\text{$\pm$}0.19}&0.35\tiny{\text{$\pm$}0.03}\\
simplewiki& 5& 0.05 & 95.92\tiny{\text{$\pm$}0.1}&98.66\tiny{\text{$\pm$}0.05}&96.39\tiny{\text{$\pm$}0.1}&43.46\tiny{\text{$\pm$}0.14}&63.77\tiny{\text{$\pm$}0.15}&43.68\tiny{\text{$\pm$}0.14}&0.22\tiny{\text{$\pm$}0.02}\\
simplewiki& 15& 0.05 & 95.88\tiny{\text{$\pm$}0.11}&98.7\tiny{\text{$\pm$}0.08}&96.33\tiny{\text{$\pm$}0.11}&43.46\tiny{\text{$\pm$}0.17}&63.77\tiny{\text{$\pm$}0.18}&43.68\tiny{\text{$\pm$}0.17}&0.21\tiny{\text{$\pm$}0.02}\\
simplewiki& 2& 0.05 & 95.91\tiny{\text{$\pm$}0.1}&98.72\tiny{\text{$\pm$}0.05}&96.34\tiny{\text{$\pm$}0.09}&43.46\tiny{\text{$\pm$}0.2}&63.85\tiny{\text{$\pm$}0.19}&43.67\tiny{\text{$\pm$}0.21}&0.21\tiny{\text{$\pm$}0.01}\\
simplewiki& 5& 0.01 & 95.81\tiny{\text{$\pm$}0.11}&98.66\tiny{\text{$\pm$}0.08}&95.93\tiny{\text{$\pm$}0.12}&43.44\tiny{\text{$\pm$}0.16}&63.84\tiny{\text{$\pm$}0.13}&43.5\tiny{\text{$\pm$}0.16}&0.06\tiny{\text{$\pm$}0.02}\\ \hline
\end{tabular}
}
\end{table}

\section{(Theoretical) Discussion of Underlying Assumptions} \label{sec:dis}
For our method to be useful, we need the refining model $f_s$ to not introduce more errors than it fixes. This can be broken down into two assumptions: (i) Predicting the second-to-last token is (theoretically) easier than predicting the next token. (ii) The inductive bias of (existing) models is better suited for second to last token prediction, i.e., the problem is not only theoretically easier but also that current models (e.g., transformers) actually perform better. \\
Let us discuss Assumption (i): While empirically we observe that the accuracy of $f_s$ is higher than that of $f_n$, this is not necessarily true for all possible datasets. We formally rely on the assumption that predicting a token $y_{t-1}$ in a bidirectional setting, i.e., ``in-between'', is easier than the next token $y_t$.
We assume that the error of an optimal predictor $p^*$ in a bidirectional setting, is less than that in a unidirectional setting, where the error is computed by comparing it to an optimal predictor $p^*$ that has access to all the context $y_{i\neq t}$ except the token to be predicted.
That is, we assume (doing a shift of indexes):
\eq{
||p^*(y_t|y_{i\neq t})-p^*(y_t|y_{0},y_1,...,y_{t-1})||>\\ ||p^*(y_t|y_{i\neq t})-p(y_{t}|y_{1},y_2,...,y_{t-1},y_{t+1})||
}

This holds true if knowing the next token $y_{t+1}$ is more useful than knowing a token $y_0$ that occurred much earlier than the token $y_t$ to be predicted (e.g., $t-1$ tokens earlier). While this does not hold for all probability distributions (i.e., all possible datasets), it holds under the assumption of locality, i.e., nearby elements matter more than distant elements. Locality is commonly found in nature, the sciences, i.e., distributed computing~\cite{bar16} and specifically, also in the context of deep learning, e.g., convolutional neural networks~\cite{sch22c}.\\
Let us discuss Assumption (ii): Intuitively, we require that irrespective of whether a problem is theoretically easy to solve or not, a model can also approximate the problem well. We can only provide conjectures as to why  next token prediction is harder than second-to-last token prediction as current explainability techniques are still emerging \cite{sch24xai}. We observe that the next token prediction model $f_n$ tends to overfit much more than the model $f_s$ having the same architecture. This suggests that the standard GPT architecture generalizes better for the second to last token prediction, while showing similar training accuracies. The most illustrative case shown in Table \ref{tab:res} is for the simplewiki dataset, where both models achieve training accuracies of 95\% to 98\%, i.e. they differ only by 3\% in training, while their validation accuracy differs by 20\%.

\section{Related Work}
A large body of works~\cite{lin22sur,sch24comes} have built on the original transformer architecture~\cite{Scale17}. The original transformer consists of an encoder and a decoder. The encoder maps input tokens to continuous representations, while the decoder uses these representations as well as inputs, along with self-attention to perform predictions such as the next token prediction in the output sequence. 
Encoder models like BERT~\cite{BERT18} and its variants~\cite{RoBER19,san19} focus on learning (universal) text representations and, therefore, include training tasks beyond next token prediction, i.e., identifying correct sentence order and predicting masked words, i.e., operating as auto-encoder to recover corrupted tokens.
In contrast, autoregressive models for text generation, commonly employ decoder-only models as witnessed by the GPT series~\cite{GPT-219,GPT-320,gpt423,gpt4o} and typically focus on next token prediction only -- with exceptions like T5, which also employs an encoder and a decoder~\cite{T520} and~\cite{XLNet19,bao20}, which predict masked tokens. Thus, while the architecture of all of these models is similar to the original transformer~\cite{Scale17,sch24comes}, they differ significantly in their purpose and training objectives.
In our work, we employ a model that predicts masked tokens like BERT using a decoder-only model like~\cite{XLNet19}. Aside from using a decoder model, we also differ from BERT in that we do not require a mask token and achieve higher training efficiency (from 15\% up to 25\% \footnote{We could go as high as 50\% for $l=2$}) by using a static setup for predictions, i.e., we only predict tokens at fixed positions, such as every $k$th token, rather than predicting tokens at random positions. Our approach is also simpler than~\cite{XLNet19}, which does not need a mask token.~\cite{XLNet19} uses attention masks to permute inputs and requires separate content and query representations, while at the same time facing challenges with conversion, effectively limiting training efficiency. That is, they only predict the last tokens in a permuted sequence.~\cite{bao20} (similarly to~\cite{T520}) aims at a unified text language model by leveraging several training objectives, e.g., autoencoding using masking as well as autoregressive modeling.~\cite{bao20} follows a similar approach as BERT, only achieving a masking ratio of 15\%.
Furthermore, models like BERT and~\cite{XLNet19,bao20,T520} typically aim to improve their learned representations or autoregressive modeling, while our refiner model $f_s$ serves more to use predictions to improve outputs of a next-token model $f_n$, i.e., our model $f_s$ only focuses on second-to-last token predictions, while $f_n$ only focuses on next token predictions.
Prior work has also attempted to improve outputs of LLMs, often using self-critique~\cite{pan24aut}. For example,~\cite{mad24} uses state-of-the-art models like GPT-4 to self-assess its outputs and improve on it iteratively.~\cite{xu24llmref} trains a feedback model. Using iterative calls to the original LLM and the feedback model tasks such as translation can be improved. While \cite{xu24llmref} relies on language output, other methods included refinements into the architecture, e.g., \cite{xie24} performs blockwise decoding to support refinements. However, claims about self-correction capabilities of LLMs have been called into question~\cite{kam24can}.
Our paper also relates to work that aims to combat hallucinations as surveyed in~\cite{ji23sur}. Many works combating hallucinations employ fundamentally different approaches; for example,~\cite{zho21} trained a model in a supervised manner on synthetic data to detect hallucinations at a token level. Our method falls into the generation-time category~\cite{pan24aut} or post-processing category~\cite{ji23sur,xia24}, as we first generate a token before potentially altering it. Many prior works were specifically trained towards detecting and correcting hallucinations based on special datasets similar to~\cite{zho21}, e.g., early works include~\cite{cao20fact,chen21imp}. One of the few generate-then-refine approaches is~\cite{dzi21}, which uses a knowledge graph to mitigate hallucinations focusing on entities like persons.

\section{Conclusions}
Transformer-based large language models have revolutionized natural language processing, with autoregressive models like GPT generating coherent text by predicting next tokens based on prior tokens, and autoencoding models like BERT capturing bidirectional context through masked token prediction.  Our ``generate-then-refine'' approach constitutes a novel method for self-correction, yielding small but consistent improvements for next token predictions by combining a standard autoregressive model with an autoencoding model that predicts the second to last token using the top-k next tokens as context.



\begin{credits}
\subsubsection{\discintname}
The authors have no competing interests to declare that are relevant to the content of this article.
\end{credits}

\bibliographystyle{splncs04}
\bibliography{refs}

\begin{thebibliography}{10}
\providecommand{\url}[1]{\texttt{#1}}
\providecommand{\urlprefix}{URL }
\providecommand{\doi}[1]{https://doi.org/#1}

\bibitem{bao20}
Bao, H., Dong, L., Wei, F., Wang, W., Yang, N., Liu, X., Wang, Y., Gao, J., Piao, S., Zhou, M., et~al.: Unilmv2: Pseudo-masked language models for unified language model pre-training. In: International conference on machine learning. pp. 642--652. PMLR (2020)

\bibitem{bar16}
Barenboim, L., Elkin, M., Pettie, S., Schneider, J.: The locality of distributed symmetry breaking. Journal of the ACM (JACM)  \textbf{63}(3),  1--45 (2016)

\bibitem{GPT-320}
Brown, T., Mann, B., Ryder, N., Subbiah, M., Kaplan, J.D., Dhariwal, P., Neelakantan, A., Shyam, P., Sastry, G., Askell, A., et~al.: {Language models are few-shot learners}. Advances in neural information processing systems  (2020)

\bibitem{cao20fact}
Cao, M., Dong, Y., Wu, J., Cheung, J.C.K.: Factual error correction for abstractive summarization models. arXiv preprint arXiv:2010.08712  (2020)

\bibitem{chen21imp}
Chen, S., Zhang, F., Sone, K., Roth, D.: Improving faithfulness in abstractive summarization with contrast candidate generation and selection. arXiv preprint arXiv:2104.09061  (2021)

\bibitem{BERT18}
Devlin, J., Chang, M.W., Lee, K., Toutanova, K.: {Bert: Pre-training of deep bidirectional transformers for language understanding}. arXiv:1810.04805  (2018)

\bibitem{dzi21}
Dziri, N., Madotto, A., Zaiane, O.R., Bose, A.J.: Neural path hunter: Reducing hallucination in dialogue systems via path grounding. In: Proceedings of the Conference on Empirical Methods in Natural Language Processing (2021)

\bibitem{ji23sur}
Ji, Z., Lee, N., Frieske, R., Yu, T., Su, D., Xu, Y., Ishii, E., Bang, Y.J., Madotto, A., Fung, P.: Survey of hallucination in natural language generation. ACM Computing Surveys  \textbf{55}(12),  1--38 (2023)

\bibitem{kam24can}
Kamoi, R., Zhang, Y., Zhang, N., Han, J., Zhang, R.: When can llms actually correct their own mistakes? a critical survey of self-correction of llms. Transactions of the Association for Computational Linguistics  \textbf{12},  1417--1440 (2024)

\bibitem{lin22sur}
Lin, T., Wang, Y., Liu, X., Qiu, X.: A survey of transformers. Open AI  \textbf{3},  111--132 (2022)

\bibitem{RoBER19}
Liu, Y., Ott, M., Goyal, N., Du, J., Joshi, M., Chen, D., Levy, O., Lewis, M., Zettlemoyer, L., Stoyanov, V.: {Roberta: A robustly optimized bert pretraining approach}. arXiv:1907.11692  (2019)

\bibitem{mad24}
Madaan, A., Tandon, N., Gupta, P., Hallinan, S., Gao, L., Wiegreffe, S., Alon, U., Dziri, N., Prabhumoye, S., Yang, Y., et~al.: Self-refine: Iterative refinement with self-feedback. Advances in Neural Information Processing Systems  \textbf{36} (2024)

\bibitem{gpt423}
OpenAI: Gpt-4 technical report  (2023)

\bibitem{gpt4o}
{OpenAI}: Hello gpt-4o! \url{https://openai.com/index/hello-gpt-4o/} (2024), accessed: 2024-09-19

\bibitem{pan24aut}
Pan, L., Saxon, M., Xu, W., Nathani, D., Wang, X., Wang, W.Y.: Automatically correcting large language models: Surveying the landscape of diverse automated correction strategies. Transactions of the Association for Computational Linguistics  \textbf{12},  484--506 (2024)

\bibitem{rad18}
Radford, A.: Improving language understanding by generative pre-training. Open AI  (2018)

\bibitem{GPT-219}
Radford, A., Wu, J., Child, R., Luan, D., Amodei, D., Sutskever, I., et~al.: {Language models are unsupervised multitask learners}. OpenAI blog  (2019)

\bibitem{T520}
Raffel, C., Shazeer, N., Roberts, A., Lee, K., Narang, S., Matena, M., Zhou, Y., Li, W., Liu, P.J.: {Exploring the limits of transfer learning with a unified text-to-text transformer}. The Journal of Machine Learning Research  (2020)

\bibitem{san19}
Sanh, V., Debut, L., Chaumond, J., Wolf, T.: {DistilBERT, a distilled version of BERT: smaller, faster, cheaper and lighter}. arXiv:1910.01108  (2019)

\bibitem{sch22c}
Schneider, J.: Correlated initialization for correlated data. Neural Processing Letters  \textbf{54}(3),  2249--2266 (2022)

\bibitem{sch24xai}
Schneider, J.: Explainable generative ai (genxai): A survey, conceptualization, and research agenda. Artificial Intelligence Review  \textbf{57}(11), ~289 (2024)

\bibitem{sch24comes}
Schneider, J.: What comes after transformers?--a selective survey connecting ideas in deep learning. arXiv preprint arXiv:2408.00386  (2024)

\bibitem{Scale17}
Vaswani, A., Shazeer, N., Parmar, N., Uszkoreit, J., Jones, L., Gomez, A.N., Kaiser, {\L}., Polosukhin, I.: {Attention is all you need}. Advances in neural information processing systems  (2017)

\bibitem{xia24}
Xiao, Z., Snoek, C.G.: Beyond model adaptation at test time: A survey. arXiv preprint arXiv:2411.03687  (2024)

\bibitem{xie24}
Xie, Y., Goyal, A., Wu, X., Yin, X., Xu, X., Kan, M.Y., Pan, L., Wang, W.Y.: Coral: Order-agnostic language modeling for efficient iterative refinement. arXiv preprint arXiv:2410.09675  (2024)

\bibitem{xu24llmref}
Xu, W., Deutsch, D., Finkelstein, M., Juraska, J., Zhang, B., Liu, Z., Wang, W.Y., Li, L., Freitag, M.: Llmrefine: Pinpointing and refining large language models via fine-grained actionable feedback. In: Findings of the Association for Computational Linguistics (2024)

\bibitem{XLNet19}
Yang, Z., Dai, Z., Yang, Y., Carbonell, J., Salakhutdinov, R.R., Le, Q.V.: {Xlnet: Generalized autoregressive pretraining for language understanding}. Advances in neural information processing systems  (2019)

\bibitem{zho21}
Zhou, C., Neubig, G., Gu, J., Diab, M., Guzm{\'a}n, F., Zettlemoyer, L., Ghazvininejad, M.: Detecting hallucinated content in conditional neural sequence generation. In: Findings of the Association for Computational Linguistics (2021)

\end{thebibliography}
\end{document}